\documentclass[letterpaper, 10 pt, conference]{ieeeconf} 
\usepackage{amsmath}
\usepackage{amsfonts}
\usepackage{graphicx}
\usepackage{siunitx}  
 
\def\xvec{\boldsymbol{x}} 
\def\uvec{\boldsymbol{u}}
\def\zvec{\hat{\boldsymbol{x}}}
\def\yvec{\boldsymbol{y}}

\def\wvec{\boldsymbol{w}}
\def\pvec{\boldsymbol{p}}
\def\pvech{\hat{\boldsymbol{p}}}

\def\vvec{\hat{\boldsymbol{u}}}
\def\rvec{\hat{\boldsymbol{y}}}
\def\evec{\boldsymbol{e}}

\def\RP{\mathbb{R}^P}
\def\R{\mathbb{R}}
\def\I{\mathbf{I}}
\def\0{\mathbf{0}}

\def\Jac{\boldsymbol{J}}

\IEEEoverridecommandlockouts

\usepackage{xcolor}

\title{\LARGE \bf
A Non-Linear Model Predictive Task-Space Controller Satisfying Shape Constraints for Tendon-Driven Continuum Robots
}

\author{
Maximillian Hachen, 
Chengnan Shentu, \textit{Student Member, IEEE,}
Sven Lilge, \textit{Member, IEEE,}  \\
and Jessica Burgner-Kahrs, \textit{Senior Member, IEEE}
\thanks{
We acknowledge the support of the Natural Sciences and Engineering Research Council of Canada (NSERC), [RGPIN-2019-04846] as well as the Canada Foundation for Innovation and Ontario Research Fund [Project \#40110]. (Corresponding author: Chengnan Shentu {\tt\small cshentu@cs.toronto.edu})
}
\thanks{
All authors are with the Continuum Robotics Laboratory, Department of Mathematical and Computational Sciences, University of Toronto, Mississauga, ON L5L~1C6, Canada
}
}

\begin{document}

\maketitle

\begin{abstract}
    Tendon-driven continuum robots (TDCRs) have the potential to be used in minimally invasive surgery and industrial inspection, where the robot must enter narrow and confined spaces. We propose a  model predictive control (MPC) approach to leverage the non-linear kinematics and redundancy of TDCRs for whole-body collision avoidance, with real-time capabilities for handling inputs at 30Hz. Key to our method's effectiveness is the integration of a nominal piecewise constant
    curvature (PCC) model for efficient computation of feasible trajectories, with a local feedback controller to handle modeling uncertainty and disturbances.
    Our experiments in simulation show that our MPC outperforms conventional Jacobian-based controller in position tracking, particularly under disturbances and user-defined shape constraints, while also allowing the incorporation of control limits. We further validate our method on a hardware prototype, showcasing its potential for enhancing the safety of teleoperation tasks.
\end{abstract}

\section{Introduction}

Tendon-driven continuum robots (TDCRs) consist of a slender and flexible backbone, along with tendons routed through spacer disks. 
Complex bending shapes can be achieved by pulling and releasing the tendons and utilizing multiple stacked segments.
TDCRs offer higher dexterity and better controllability compared to other design principles for CRs \cite{Zheng2017}.
As a result, TDCRs are proposed for inspection and repair tasks in jet engines \cite{Dong2017, Wang2021}, aiming to save time and cost associated with disassembly.
In the medical field, TDCRs are proposed for surgery to facilitate minimally invasive procedures \cite{Sadati2015}.
In both cases, TDCRs are teleoperated in highly confined and unstructured environments, necessitating insertion into closed environments through small incisions or entry ports.

Contacts with the environment are often considered tolerable in CR research as the robot's structure is compliant and thus inherently safe.
However, if the TDCR has increased stiffness \cite{Rao15} or if the environment is considered fragile (e.g. near blood vessels), this assumption no longer holds and collisions must be actively avoided.
Consequently, task-space controllers or path-planners are needed to steer the tip of an TDCR while respecting environmental constraints to avoid collisions with the robot's body.
Teleoperation further requires such methods to be real-time capable as target way points are not known before deployment.

\begin{figure}
    \centering
    \includegraphics[width=0.7\columnwidth]{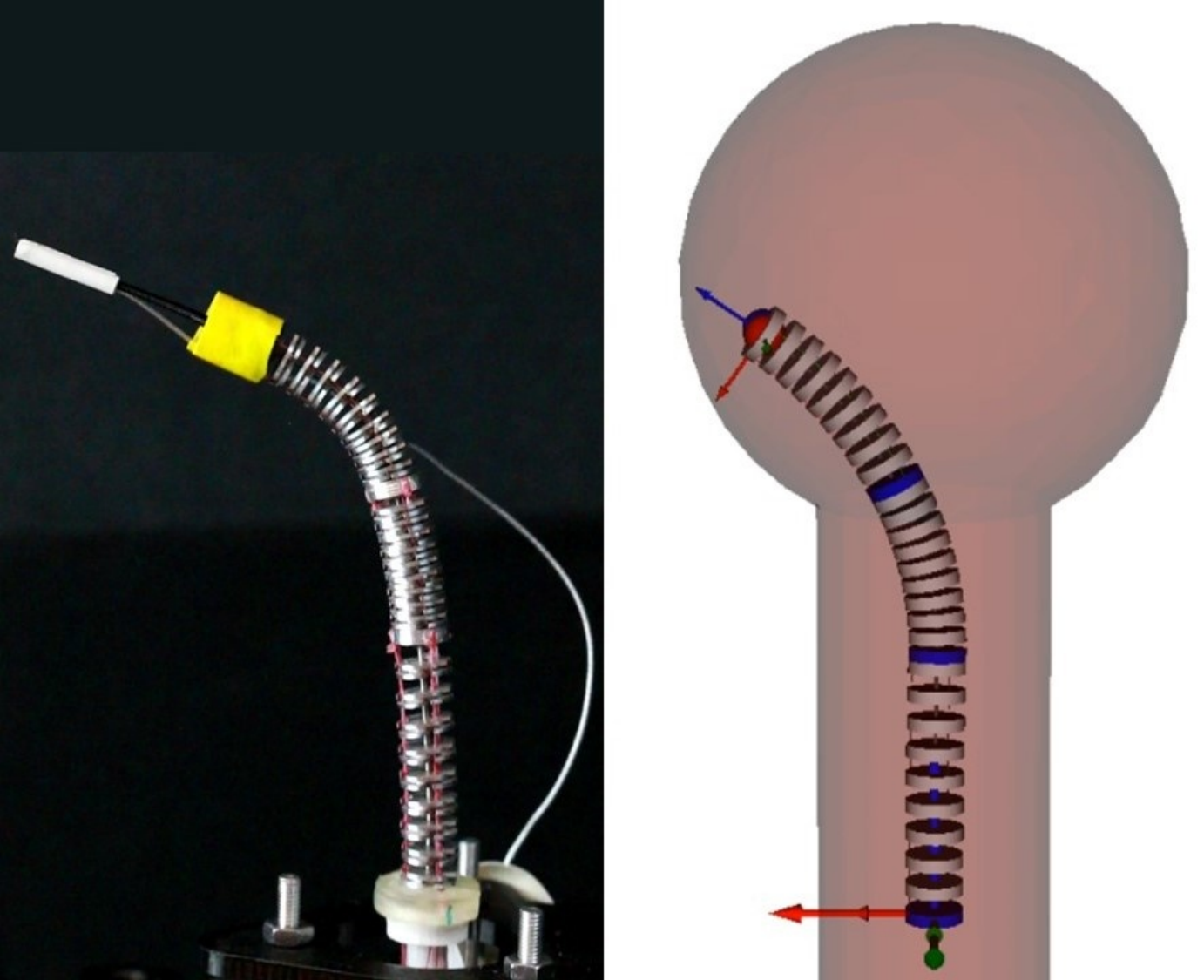}
    \caption{
        Our proposed MPC controller enables TDCRs (left) to navigate between target end-effector positions while avoiding collision with a user-defined safe zone (right), which is crucial for safe teleoperation.
        Please see attachment for videos.
    }
    \label{fig:catchy}
\end{figure}
\vspace{-1mm}
\subsection{Related Work}

Various task-space controllers have been proposed for TDCRs, which can be broadly categorized into model-free and model-based.

\subsubsection{Model-Free Approaches} 
In \cite{Back2017}, a model-free position controller was proposed for a tendon-driven catheter, where the tendon tensions are controlled directly based on tip pose. Although the formulation is efficient, it is limited to one bending segment.
A model-less feedback controller was proposed in \cite{Yip2014}, which estimates the robot's Jacobian from end-effector (EE) position feedback and actuator inputs at every iteration, and then used to solve for tendon tension to reach a desired EE position.
Other model-free approaches leverage learning-based techniques \cite{Wang2021a}.
For instance, in \cite{Abdulhafiz2021}, a deep-learning network directly maps camera images to motor commands, achieving accurate positioning even in challenging lighting conditions.
In \cite{Lai2023}, reinforcement learning on a real robot is used to obtain a control policy, with a proposed axis-space framework to reduce training time and improve accuracy.
While model-free approaches exhibit robust handling of obstacles and disturbances,
explicit formulation of obstacle avoidance constraints remains a challenge.

\subsubsection{Model-Based Approaches} 

Path following motions of a TDCR with extensible segments was realized by computing joint-space trajectories using a sequential quadratic programming scheme and a Cosserat rod model in \cite{Amanov2021}.
While this strategy enables complex pre-computed follow-the-leader motions, potentially also in cluttered environments, it is executed open-loop and does not allow for online adaption to obstacles.

In \cite{Roesthuis2016}, a rigid-link model for TDCR was proposed. A pseudoinverse Jacobian controller, identical to algorithms for serial robots \cite{Deo95}, is used for trajectory tracking.
Obstacle avoidance is achieved through null-space joint motions, but explicit hard-constraints are not considered.
The framework is evaluated on a two-segment planar TDCR with integrated shape sensing, and achieved good tracking result while avoiding static and moving circular obstacles.

In \cite{Norouzi2021}, a model predictive control (MPC) controller with visual servoing is used for tip tracking of a 3-DOF tendon-driven catheter.
The MPC controller is formulated with a PCC model, and is subject to actuator velocity and visibility constraints.

In \cite{Seleem2023}, a dynamic robot model was computed using a Lagrangian formulation based on the PCC model of a two-section TDCR to build an impedance controller. On top of low level control, an imitation based motion planner was developed for planning trajectories while avoiding small dynamic and static obstacles (1-\SI{2}{cm}). The algorithm excels at repeatable movements but suffers from oscillatory behaviour with volumetric obstacles and slow computation time.

\cite{Lai2022} derived the curvature kinematics of a cable-driven soft robot and used it to formulate a damped least squares (DLS) problem taking into account fixed end-effector orientation and obstacle avoidance. The optimization was computed iteratively offline achieving obstacle avoidance and a tracking error within \SI{5}{mm}.

Model-based approaches reveal a trade-off between accuracy and computation time. The PCC assumption is commonly used due to its low computation time and minimal state feedback requirements. 
However, limitations arising from unmodeled effects and hardware discrepancies should be accounted for in controller design to reduce steady-state errors and improve stability. 
Furthermore, while safety constraints and collision avoidance terms are considered in some works \cite{Roesthuis2016, Yip2014, Seleem2023}, they are limited to simple convex obstacles like points and spheres. 
This setting does not adequately capture the real-world use-cases for TDCR, such as insertion into confined operation areas. 

\subsection{Contribution}

This work aims to provide a task-space control algorithm for TDCRs capable of avoiding body collisions in narrow environments. The main contributions of this letter include: 1) A MPC formulation based on the PCC assumption, balancing efficiency with robustness against unmodeled effects;
2) Explicit handling of arbitrary obstacles via MPC constraint formulation;
3) Comprehensive evaluation of its performance in simulation and hardware experiments, offering insights on the limitations of Jacobian-based controllers for tasks with body-collision constraints.

\section{MPC Formulation}
\begin{figure}[h]
	\centering
    \includegraphics[width=\columnwidth]{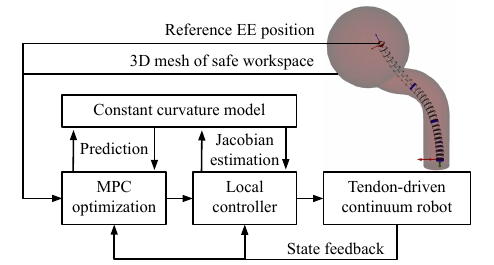}
    \caption{
        We integrate a nominal MPC for tip tracking and collision avoidance, with a local controller for disturbance rejection. The PCC model provides efficency for real-time performance, while the hierarchical architecture provides robustness against modeling uncertainty and disturbances.
    }
    \label{fig:systemOverview}
\end{figure}

Due to the elastic nature of TDCRs, finite state representations inevitably introduce modeling uncertainties in addition to physical disturbances affecting the system. 
To address this, we propose a hierarchical controller formulation, shown in Fig.~\ref{fig:systemOverview}. We integrate a MPC utilizing the PCC model for efficient real-time computation of feasible trajectories, with a local controller designed to address modeling errors and disturbances as they occur and to maintain the disturbed real system state close to the nominal state.

\subsection{Robot State and Output}
Throughout this work, we consider a three-segment tendon-driven continuum robot with extensible segments, operated by three tendon actuators and one backbone actuator per segment as described in \cite{Amanov2021}. We assume that tendon routing disks are equidistantly spaced per segment and that the number $n$ of disks per segment is identical.
The robot state vector $\xvec \in \mathbb{R}^P$ with the total number of actuators $P=12$ is formulated as:
\begin{align}
	\xvec = \begin{pmatrix}
		q_{1,1} &
		q_{1,2} &
		q_{1,3} &
		\gamma_1 &
		q_{2,1} &
		q_{2,2} &
        \cdots &
	\end{pmatrix}^T
	\label{eq:xdef}
\end{align}
where $q_{j,m}$ denotes the absolute tendon length and $\gamma_j$ the segment length, with segment index $j\in\left\{1,2,3\right\}$ and tendon index $m\in\left\{1,2,3\right\}$.

\begin{figure}[ht]
    \centering
    \includegraphics[width=0.5\columnwidth]{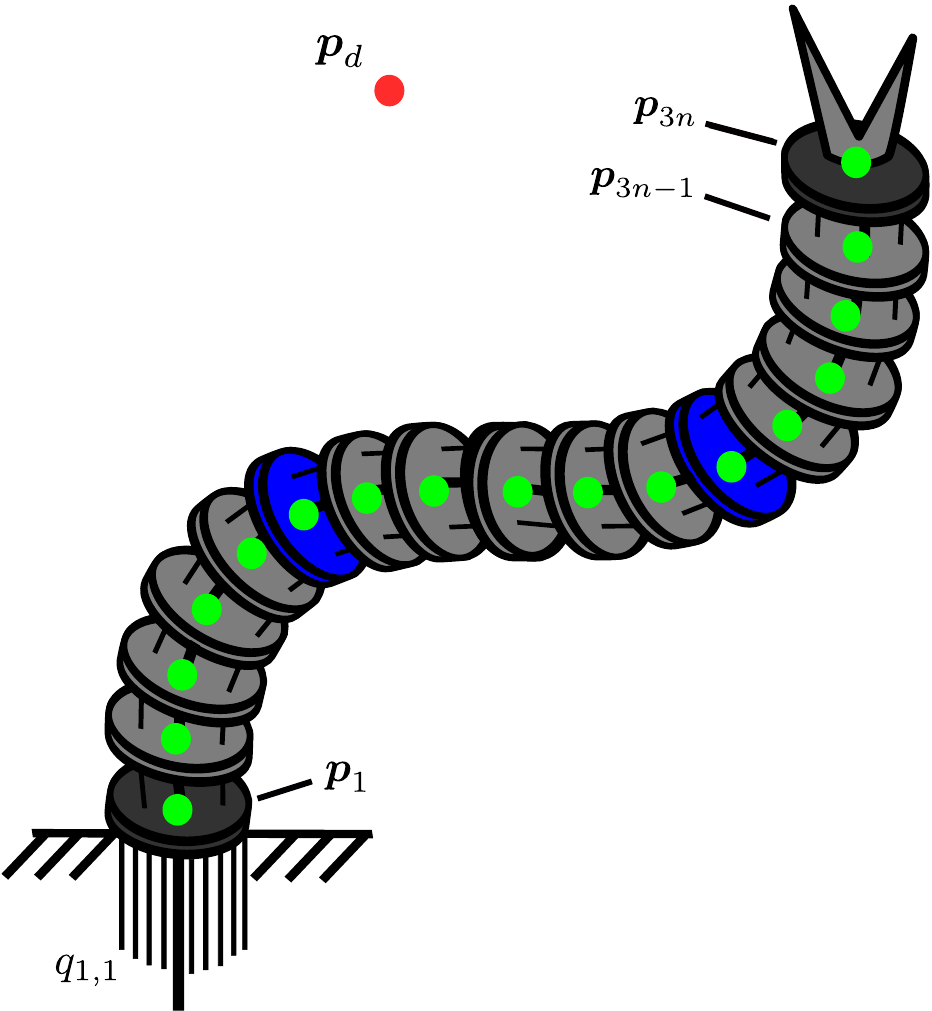}
    \caption{
        A three-segment TDCR. Blue coloured tendon-routing disks mark the end of each segment. The robot state $\xvec$ is composed of the actuator variables $q_i,m$ and $\gamma_m$. The output shape $\yvec$ is composed of the tendon-routing disk positions $\pvec_1$, \dots, $\pvec_{3n}$. The position $\pvec_d$ is the desired end-effector position.   
    }
    \label{fig:variables}
\end{figure}

A key requirement for choosing a model for MPC is computational efficiency, given that the model needs to be solved several times per prediction and optimization iteration. 
Therefore, we use a PCC model as it is the computationally fastest method for modelling load-free TDCR kinematics \cite{Rao15}. 
The PCC forward kinematics model $f_{PCC}:\RP \rightarrow \mathbb{R}^{3n}$ provides a mapping from robot state $\xvec$ to its shape $\yvec$:
\begin{align}
	\yvec = \begin{pmatrix}
		{\mathbf{p}_1}^T&
		{\mathbf{p}_2}^T&
		\hdots &
		{\mathbf{p}_{3n}}^T
	\end{pmatrix}^T
\in \mathbb{R}^{3n},
\end{align}
where $\mathbf{p}_i\in \mathbb{R}^{3}$ with $i\in\left\{1,\dots,3n\right\}$ is the position of the respective tendon routing disk.
The target end-effector position is defined as $\pvec_d \in \mathbb{R}^{3}$.
A visualization of all robot state and output variables is shown in Fig.~\ref{fig:variables}.

\subsection{Time Dependent Modelling}
\label{subsec:timeModel}
To derive a time-dependent state space model, we consider the robot movements to be slow, such that dynamic effects can be neglected (i.e. quasi-static), which is a common assumption in continuum robots \cite{Amanov2021, Norouzi2021, Li2018}.
The time index $\left[k\right]$ is chosen with a time step of $\Delta t$ dependent on the controller frequency.
With the change in actuator length per time step:
\begin{align}
\uvec[k] = \frac{\Delta \xvec[k]}{\Delta t} = \frac{\xvec[k] - \xvec[k-1]}{\Delta t} \label{}
\end{align}
i.e. the actuator speed, as the control input, the state space model follows an integrator:
\begin{align}
	\xvec[k] =  \xvec[k-1] + \Delta t \uvec[k]
    .
    \label{systemdyn}
\end{align} 

\begin{figure}
	\centering
    \includegraphics[width=\columnwidth]{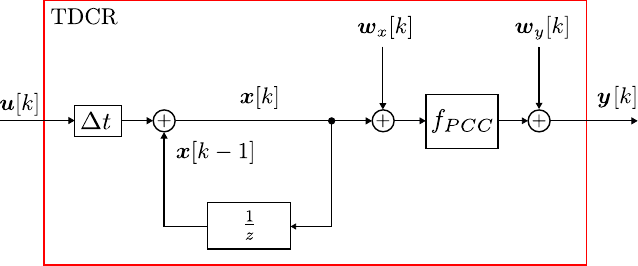}
    \caption{
        The TDCR's joint spcae is used as its state, with a non-linear and disturbed output function based on the PCC model.
    }
    \label{fig:systemModel}
\end{figure}

The PCC model is a kinematic model and neglects external forces.
To account for errors induced by such forces acting on the robot, we introduce two disturbance terms:
\begin{align}
    \wvec_x \sim \mathcal{N}(\boldsymbol{0}, {\sigma_x}\mathbf{I})  \text{ with} - w_{x,max} \leq \wvec_{x,i} \leq w_{x,max}\\
     \wvec_y \sim \mathcal{N}(\boldsymbol{0}, {\sigma_y}\mathbf{I})  \text{ with} - w_{y,max} \leq \wvec_{y,i} \leq w_{y,max},
\end{align}
where $w_{x,max},w_{x,max},{\sigma_x},{\sigma_y} > 0 $ and the identity matrix $\mathbf{I}$ is with dimension $3n \times 3n$.
The disturbance terms are included in the system description by adding them to the output function:
\begin{align}
    \label{stateModel}
	\yvec[k] = f_{PCC}(\xvec[k]+\wvec_x[k]) + \wvec_y[k]. 
\end{align}
A block diagram of the system model is shown in Fig.~\ref{fig:systemModel}. 
The proposed system model is a linear integrator in state space with a non-linear and disturbed output function.

\subsection{Model Predictive Controller}
 
As no disturbance terms are considered by the nominal MPC, the predicted states, inputs and outputs can deviate from the actual ones of the system (denoted by $\xvec$, $\uvec$ and $\yvec$).
To distinguish between predicted and measured entities, we introduce the predicted state $\zvec$, the predicted input $\vvec$ and the predicted output $\rvec$.
The system description is then:
\begin{align}
		\zvec&[k] =  \zvec[k-1] + \Delta t\vvec[k]\\
  \label{nominalkinematics}
		\rvec&[k] = f_{CC}(\zvec[k]).
\end{align}
Future trajectories predicted at time $k$ are denoted with the index $[i|k]$ with $i =  0,\dots,N$, with $N\in \mathbb{N}^0$ being the prediction horizon at each iteration.
\subsubsection{MPC Cost Function}
For the EE to converge towards $\pvec_d$, 
we define a stage cost $L$ at every predicted time step $i$ at time $k$, omitting the index $[i|k]$ for brevity:
\begin{align}
	&L(\zvec,\vvec) = \mathbf{\evec}_{EE}^T \mathbf{Q} \mathbf{\evec}_{EE} + \vvec^T          \bf{R} \vvec + \zvec^T \bf{S} \zvec \label{L},\\
	&\text{with: } \nonumber \\
	&\evec_{EE} = \pvech_{3n} - \pvec_d \label{ez}.
\end{align}
Thus, $L$ is defined as a quadratic cost over the nominal EE error $\evec_{EE}$ (\ref{ez}), the nominal input $\vvec$ and the nominal state $\zvec$.
For the weighting matrices $\bf{Q}$, $\bf{R}$ and $\bf{S}$, time-invariant weightings are chosen.
By \eqref{ez}, the nonlinear forward kinematics of the TDCR are indirectly considered as $\rvec$ is computed with \eqref{nominalkinematics}.
The first term in \eqref{L} is chosen such that the EE position converges to $\pvec_d$.
The second term, $\vvec^T \bf{R} \vvec$, minimizes the used actuator effort for reaching the goal. 
The third term, $\zvec^T \bf{S} \zvec$, reduces the absolute length of tendons and segments resulting in robot configurations with lower curvatures.
\subsubsection{The MPC Optimization Problem}

The MPC optimization problem is then:
\begin{align}
	&\underset{\zvec[\cdotp|k], \vvec[\cdotp|k]}{\operatorname{min}}  
\sum_{i=0}^{N-1} L(\zvec[i | k], \vvec[i | k]),\\[1em]
	\text {subject to: }& \nonumber \\
	& \zvec[0 | k]=\zvec_0[k] \\
	& \text{for } i=0,1, \ldots, N-1 :\nonumber \\
	& \zvec[i+1 | k] = \zvec[i | k] + \Delta t\vvec[i + 1 | k]\\
	& \zvec[i | k] \in \mathbb{X} \\
	& \vvec[i | k] \in \mathbb{U}.
\end{align}
We define here, the state constraint set $\mathbb{X} \subseteq \RP$, the input constraint set $\mathbb{U} \subseteq \RP$, and the initial state $\zvec_0 \in \mathbb{X}$.
In addition, the cost function $J_N$ is defined as the summation of the stage cost function $L: \RP \times \RP \rightarrow \R$ over the prediction horizon $N \in \mathbb{N}^0$. %
The optimal predicted state trajectory, resulting from solving the nominal optimization problem is denoted as $\zvec^*[\cdot | k]$, the optimal predicted input trajectory as  $\vvec^*[\cdot | k]$ and the optimal output with $\rvec^*[\cdot | k]$.

\subsubsection{Initializing the MPC Predictions}

The initial state $\zvec_0$ at each time step must be feasible.
The initial state at the next time step is defined as the first optimal predicted state of the last prediction, namely:
\begin{align}
	\zvec_0[k] = \zvec^*[0 | k-1].
\end{align}
This is feasible as the disturbance in the system model is only acting additive on the output (see \eqref{stateModel}).
We consider the nominal state at the very first iteration to be known as a straight configuration with all segments at the same length in the center of the joint limits.

\subsubsection{MPC Constraints}

The primary control task, reaching the desired EE position, is achieved with the MPC cost function.
The secondary control task, avoiding collisions with the environment, is formulated as a MPC constraint.

We allow general shape constraints in the task space, represented by a 3D mesh, as a user-defined safe zone.
Based on the 3D mesh, we introduce an MPC constraint that restricts points described by the nominal output vector $\rvec$ within this safe zone.
We can define a signed distance function $d: \R^3 \rightarrow \R$ with respect to the input 3D mesh, where
$d$ is positive if the query position is contained in the mesh, and negative if outside \cite{Barentzen2005}.
The constraint for the robot's shape is then:
\begin{align}
    \label{hullconst}
	d(\pvec_i[k]) > c_d\text{ for } i=1,\dots,3n
\end{align}
with a safe distance margin $c_d > 0$.
As $\pvec_i[k]$ is part of $\rvec$, which is a function of the state $\zvec[k]$, this is a non-linear constraint on the robot state.

Besides the shape constraint for collision avoidance, there are straightforward limits on tendon length, speed and segment curvature:
For tendon speed and length limitation, the feasible actuator velocities and states are restricted by simple box constraints.

Next, there is a mechanical constraint inherent to tendon actuation, that tendons on one side have to be released, if tendons on the other side are pulled, expressed as:
\begin{align}
     \label{qcond}
	 3 \gamma_j - \sum_{m=1}^{3} q_{j,m} = 0 \quad\forall j\in\left\{1,2,3\right\}
\end{align}
with the segment length $\gamma_j$ and absolute tendon length $ q_{j,m}$ being part of the predicted state vector $\zvec$.

The curvature of each segment may need to be limited to prevent damage to the robot.
The maximum allowed curvature and the current curvature for a segment $j$ can be described as a continuous non-linear function $\kappa_{max,j}(\gamma_j)$ and $\kappa_j(q_{j,1},...,q_{j,3},\gamma_j)$. 
With these, the curvature of each segment can be constrained to the corresponding maximum curvature:
\begin{align}
\label{curvmax}
	\kappa_j(q_{j,1},...,q_{j,L},s_j) < \kappa_{max,j}(s_j) \text{ }\forall j\in\left\{1,2,3\right\}.
\end{align}

\subsection{Local Controller}
For dealing with the disturbances $\wvec_x[k]$ and $\wvec_y[k]$ introduced in Sec.~\ref{subsec:timeModel}, an analytic local control law $K_{loc}:\R^n \rightarrow \R^P$ is used.
This controller should steer the real system output $\yvec[k]$ in proximity to the optimal nominal output $\rvec^*[k]$.

As a control scheme, we use the well-known Pseudo-Inverse Jacobian approach and apply it to a TDCR, similar to \cite{Roesthuis2016}.
As we seek to control the entire robot body, the Jacobian of not only the EE position, but at all disk positions along the robot, are considered.
We numerically approximate these Jacobians by a finite difference approach and denote them as $\Jac_i$ with $i=1,...,3n$, where
$\Jac_0$ is at the fixed base disk, $\Jac_1$ at the first disk, and $\Jac_{3n}$ is the EE (denoted by $\Jac$ omitting $3n$ in the following).

The equations for the local controller can now be expressed as:
\begin{align}
	\label{Kloc}
	&K_{loc}(\rvec^* - \yvec) = K_{EE} + K_{body}
\end{align}
with
\begin{align}
	K_{EE} &= k_{EE} \Jac^+ \boldsymbol{\Delta}_{EE} \label{KlocEE}\\
	\Jac^+ &= (\Jac^T \Jac + \lambda \I_{P \times P})^{-1} \Jac^T \text{,    }\lambda > 0\\
	\boldsymbol{\Delta}_{EE} &= \pvech_{3n} - \pvec_{3n} \label{deltaee}
\end{align}
and
\begin{align}
     K_{body} &= k_{body} (\I_{P \times P} - \Jac^+ \Jac) \frac{\partial \boldsymbol{\Delta}_{body}}{\partial \zvec} \label{KlocBody}\\
	\boldsymbol{\Delta}_{body} &= \frac{1}{3n-1} \sum_{i=1}^{3n-1} (\pvech_{i} - \pvec_{i})^T(\pvech_{i} - \pvec_{i}) \label{deltabody}\\
     \frac{\partial \boldsymbol{\Delta}_{body}}{\partial \zvec} &= \frac{2}{3n-1} \sum_{i=1}^{3n-1} (\pvech_i - \pvec_i)^T \Jac_i \label{deltabodydx}\\
    \Jac_i &= \frac{\partial \pvec_i}{\partial \zvec}\label{Jacobian_i}.
\end{align}
We note that these equations omit the index $[k]$ for brevity.
The control law consists of two terms: \eqref{KlocEE} for steering the EE and \eqref{KlocBody} for moving the robot body.
Both terms feature a constant weighting factor, $k_{EE}$ and $k_{body}$ respectively, the  Pseudo Inverse Jacobian $\Jac^+$ and an error term.
Here, $\boldsymbol{\Delta}_{EE}$ \eqref{deltaee} describes the EE position error between nominal and real state.
$\boldsymbol{\Delta}_{body}$ \eqref{deltabody} is an average squared distance function over all spacer disks, excluding the tip. Note that $\pvec_i $ and $\pvech_i$ are part of the output vectors $\yvec$ and $\rvec$ respectively. $\Jac_i$ denotes the robot Jacobian at the point $\pvec_i$ along the robot body.
Applying the method, the tip error $\boldsymbol{\Delta}_{EE}$ converges to zero and the shape error $\boldsymbol{\Delta}_{body}$ is minimized \cite{Deo95}.

\subsection{Complete Controller}

The control input to the system now follows as the summation of the MPC input and the input of the local controller, namely
\begin{align}
	&\uvec[k] =  \vvec^*[1|k] + K_{loc}(\rvec^*[k] - \yvec[k]).\label{controlInput}
\end{align}

\section{Evaluation in Simulation}

We evaluate the performance of the proposed controller in simulation and compare it with a reference controller based on damped least-squares (DLS) \cite{Deo95} modified with an additional collision avoidance cost function. 
The tests evaluate the convergence and performance of EE position tracking under disturbances and collision avoidance constraints.

\subsection{Reference Controller Formulation}

A DLS controller with an additional collision avoidance cost is implemented as a reference for comparison.
The controller, using the same notation as before, is defined as:
\begin{align}
	\uvec=k_j \left(\boldsymbol J^T \boldsymbol J+\boldsymbol W \right)^{-1}\left(\boldsymbol J^T (\pvec_d - \pvec_{3n}) + \boldsymbol W \frac{\partial h}{\partial \boldsymbol{u}} \right)
\end{align}
with $\boldsymbol{W}=c_w \I_{P \times P}$, and $k_j>0$.
In the tests, the parameters were tuned to: $c_w = 1$ and $k_j = 0.05$.
To formulate the additional collision avoidance cost $h$, the signed distance function $d$ from the MPC is used to steer the robot body into the user-defined safe zone.
We define $h$ as:
\begin{align}
	h = \sum_{i = 1}^{3n-1} (ReLU(c_d - d(\pvec_i)))^2.
\end{align}
By using an activation function, the value of the cost function is zero within the hull.
The derivative of the function is therefore also strictly zero inside and does not influence the control law. 
If the signed distance is smaller than the safe distance margin $c_d$, the cost increases quadratically.

\subsection{Implementation Details}

In terms of MPC parameters, the prediction horizon length is chosen as $N=2$ as this is the largest prediction horizon achieving an update rate of \SI{30}{Hz} with the used hardware.
Since the MPC is intended for teleoperated applications, real time update rate is prioritized over long prediction horizon.
We further choose the weighting matrices to be: $\mathbf{Q} = 1000\mathbf{I}_{9n\times 9n}$, and $\mathbf{R} = \mathbf{S} = 10\mathbf{I}_{P\times P}$.
During simulations, the disturbances $\wvec_x$ and $\wvec_y$ were drawn from a normal distribution at a lower frequency (\SI{5}{Hz}), with  standard deviations $\sigma_x = \SI{0.2}{mm}$, $\sigma_y = \SI{1}{mm}$, and maximum values $\wvec_{x,max} = \SI{2}{mm}$ and $\wvec_{y,max} = \SI{5}{mm}$.

The EE error for the real system $\evec_{EE,real}$ is defined as the Euclidean distance between EE tip and goal position.
In the nominal case, the EE error is calculated the same way for the nominal system's tip and is named $\evec_{EE,nom}$.
To show the influence of the local controller, the Euclidean error between nominal system output $\rvec^*$ and simulated output $\yvec_{r}$ is denoted as $\evec_{EE,local}$.
Furthermore, the body error as the average point-wise error over all measured points along the robot body is:
\begin{align}
	\evec_{body,local} = \frac{1}{3n-1} \sum_{i=1}^{3n-1} \| \pvec^*_{i} - \pvec_{i} \|_2.
\end{align}
To visualize a measure for the effort of the controller, the Euclidean norm of the input vector $\uvec$ is used.
In the MPC case, the Euclidean norms of both the nominal input $\vvec^*$ and the local controller input $K_{loc}(\rvec^* - \yvec)$ are used.

\subsection{Convergence Under Disturbances}

We verify the convergence of the MPC controller with a series of sampled positions within the workspace as step input, and compare the system response against the reference controller.
Collision avoidance constraints are not enforced.

\begin{figure}[t]
	\centering
    \includegraphics[width=\columnwidth]{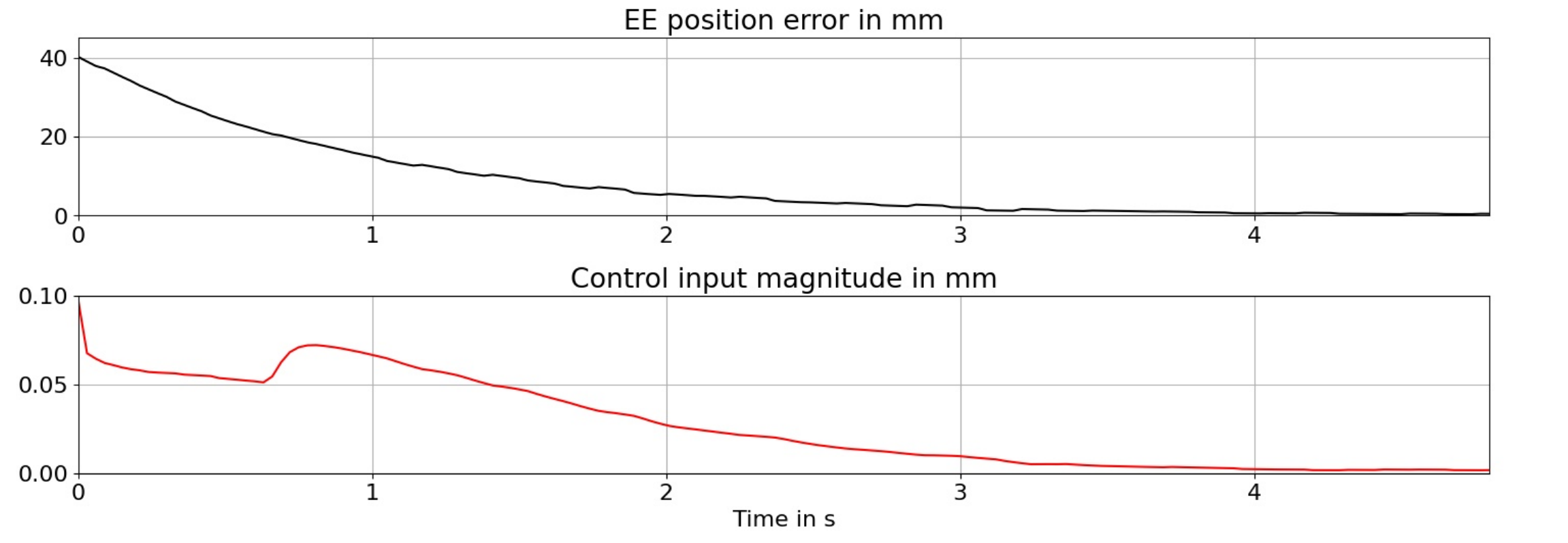}
    \caption{
        Step response of the reference DLS controller for reaching a target position \SI{40}{mm} away with disturbances. 
    }
    \label{fig:Jac_Robsut_3}
\end{figure}

\begin{figure}[t]
	\centering
    \includegraphics[width=\columnwidth]{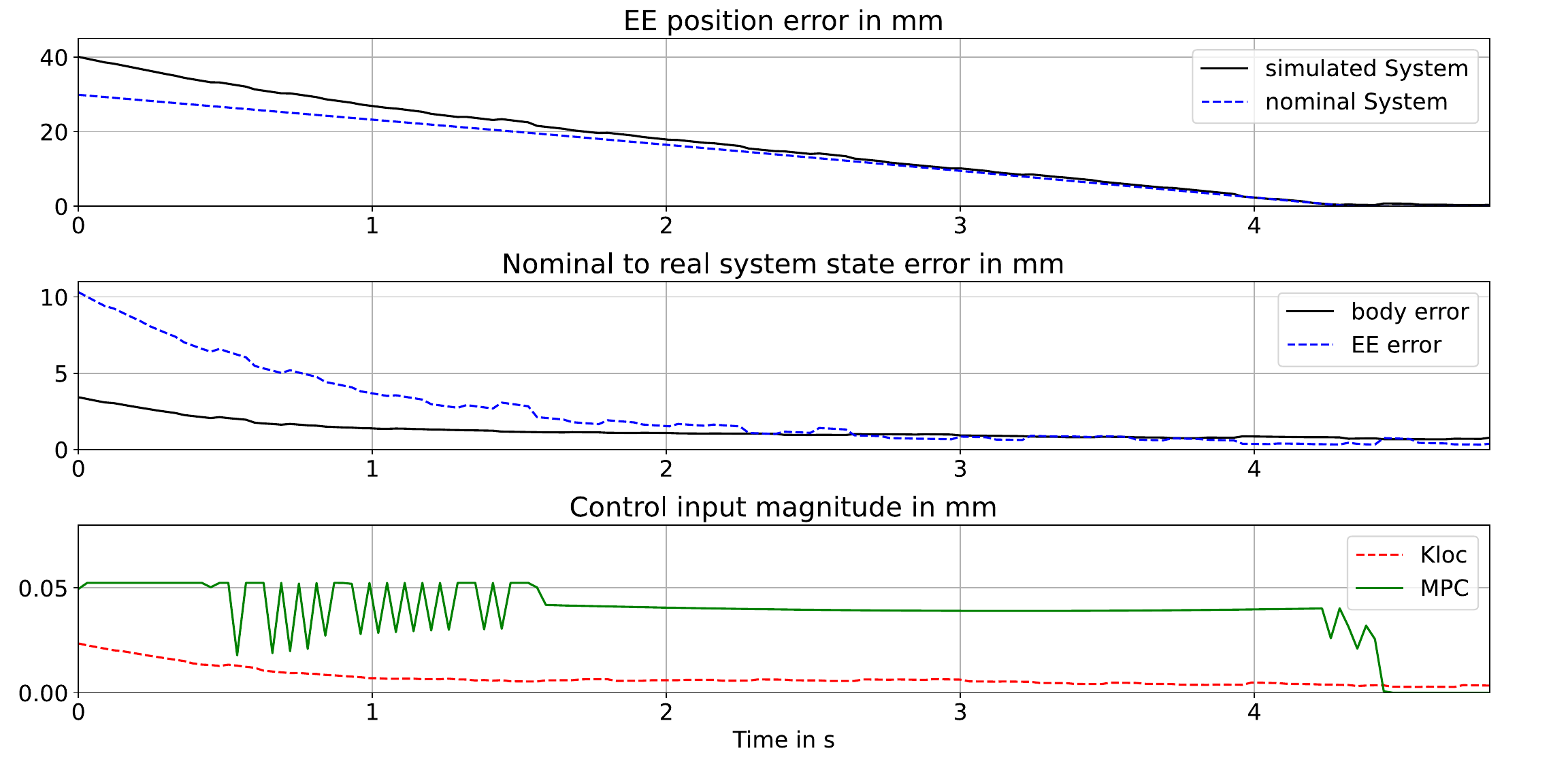}
    \caption{
        Step response of the MPC controller for reaching a target position \SI{40}{mm} away with disturbances.
        }
    \label{fig:N_1_Robust_3}
\end{figure}

By considering a normal-distributed and low-frequent disturbance, we can simulate effects like friction on the tendons and slow acting external forces.
Results from reaching one of the target positions are presented in Fig.~\ref{fig:Jac_Robsut_3} and Fig.~\ref{fig:N_1_Robust_3}, for the reference controller and the MPC respectively.

Both controllers show convergence to the desired set point within approximately the same amount of time, demonstrating their robustness against potential external disturbances and deviations from the PCC assumption.
A sudden decrease in input magnitude (\SI{0}{s} to \SI{0.75}{s} in Fig.~\ref{fig:Jac_Robsut_3}) is visible which can be explained by the kinematics of the robot.
When the decrease occurs, the robot is in a configuration where moving the first robot segment results in significant decrease of the EE error. 
As the modified DLS controller is designed to find a compromise between decrease in EE error and actuator velocities \cite{Deo95}, lowering the input magnitude while still maintaining an exponential decrease in EE error promises the most efficient movement for the controller.

The MPC controller shows a different behaviour as the nominal system error converges approximately linear to zero.
This can be explained by the input constraints, which limit the maximal EE speed. 
As the MPC optimizer tries to reach the target position as fast as possible within the given constraints, the input limitations are used to their full extend.
As dynamic effects are not modelled and accelerations not limited, the controller does not account for acceleration limits, and drives the system in a bang-bang control way.
In the MPC input, slight oscillations occur which reduce when increasing the prediction horizon.
The local error $\evec_{EE,local}$ converges to zero as expected.
The body error $\evec_{body,local}$ converges but has at a steady state error around \SI{1}{mm}.
This is likely due to the errors originating from the PCC model. Specifically, the local controller cannot exactly drive the robot to PCC shapes due to limited available degrees of freedom for actuation. In this case, the local controller cannot fully reject the output disturbances $\wvec_y$ since it does not align with the PCC assumption. This result highlights the trade-off between computation and model fidelity for control, as well as the inherent under-actuation property of TDCRs.

\subsection{Convergence Under Shape Constraints}

\begin{figure}[h]
    \centering
    \includegraphics[width=0.23\columnwidth]{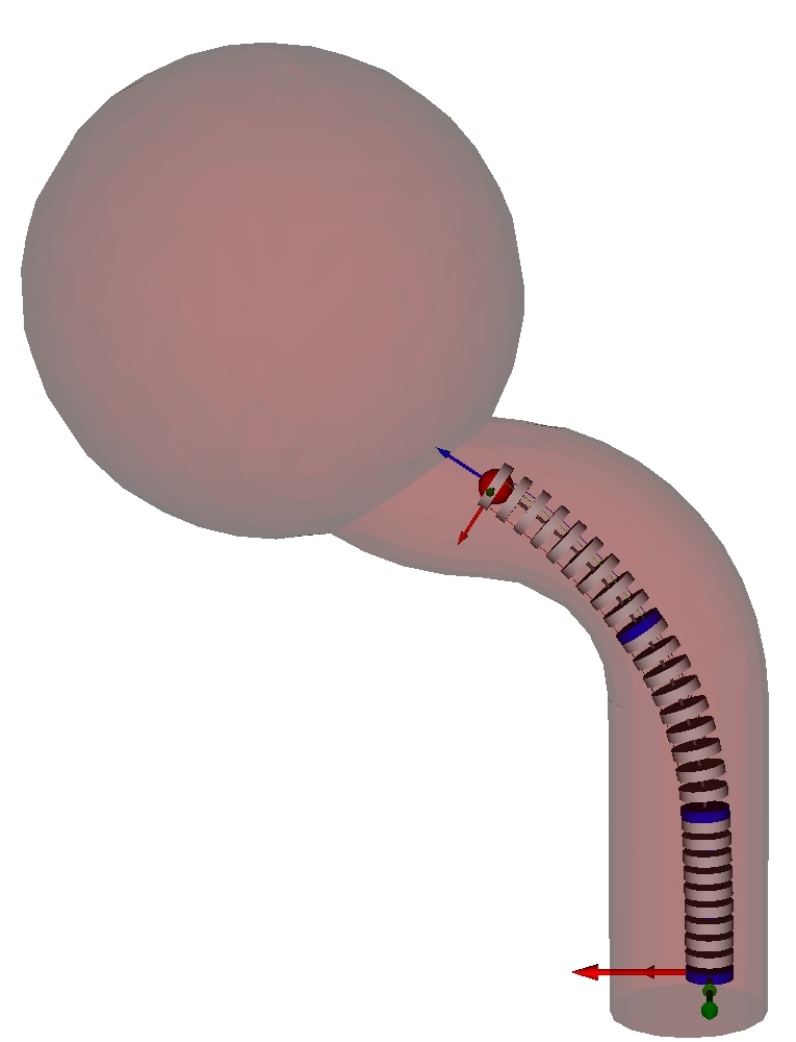}
    \hfill
    \includegraphics[width=0.23\columnwidth]{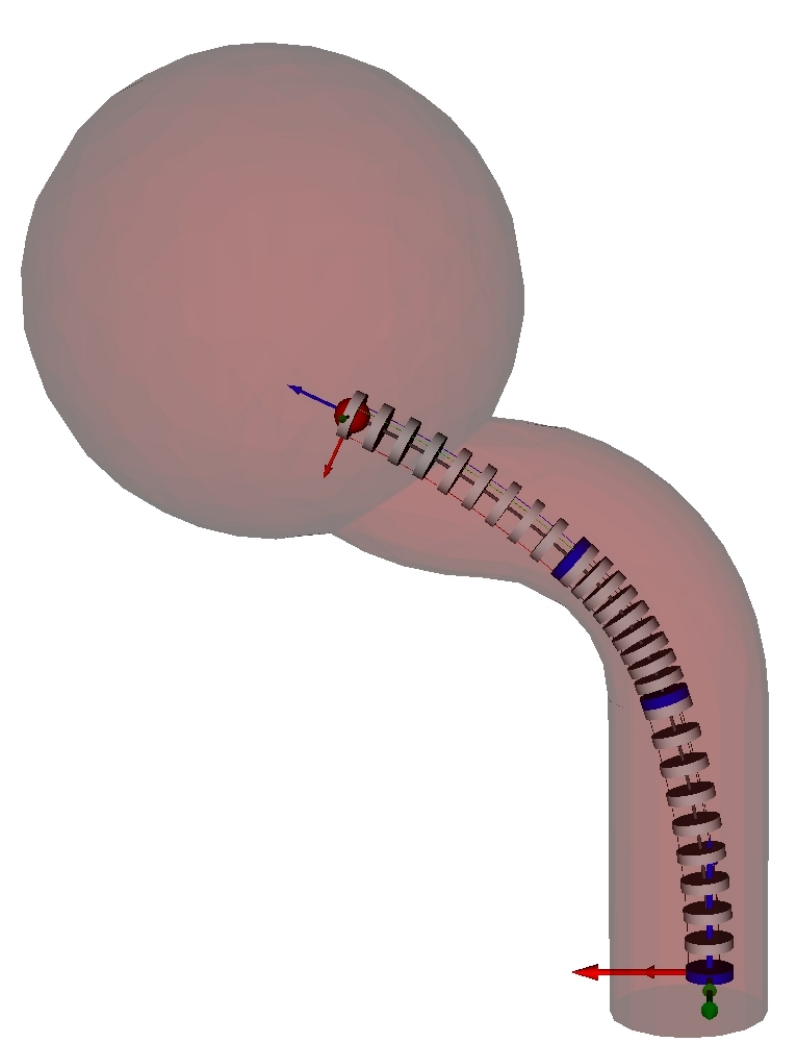}
    \hfill
    \includegraphics[width=0.23\columnwidth]{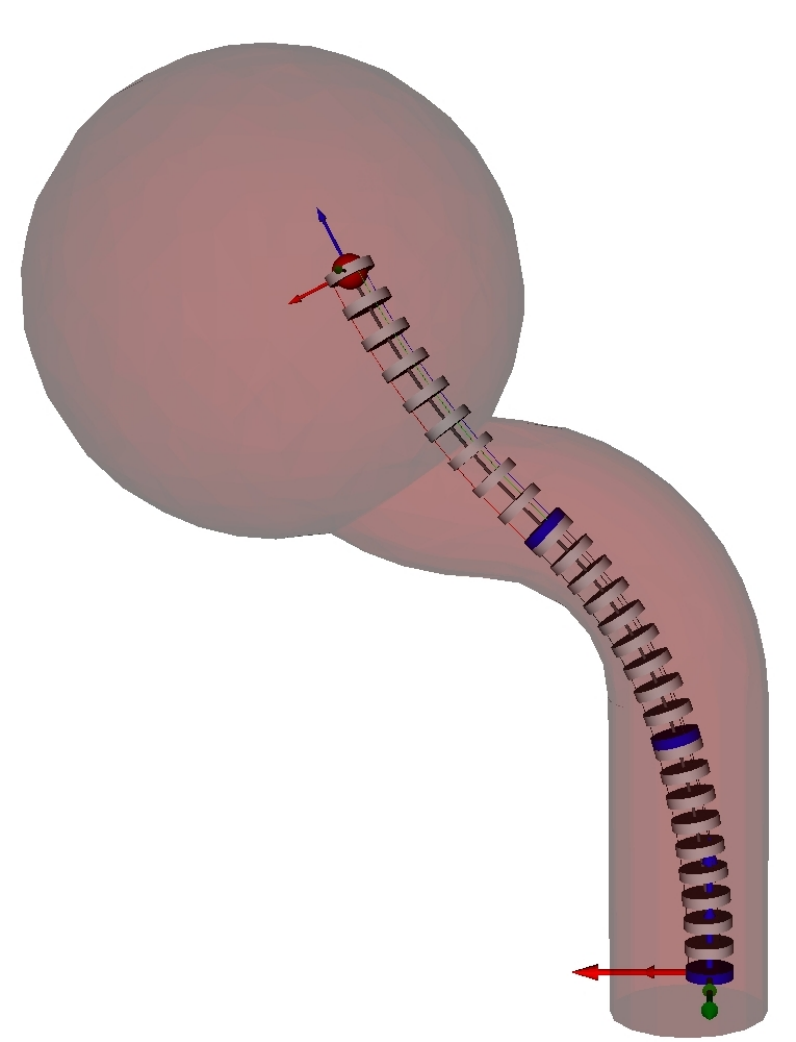}
    \hfill
    \includegraphics[width=0.23\columnwidth]{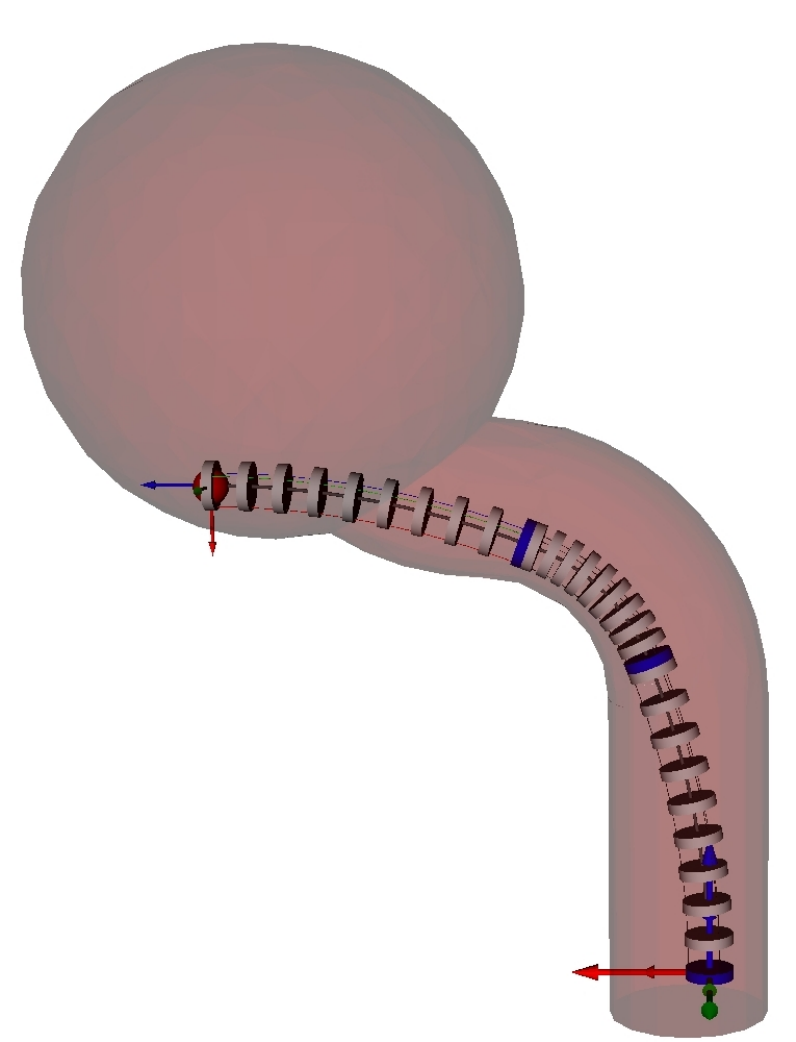}
	\caption{
        Trajectory resulting from application of the MPC controller closed loop steering to four consecutive points describing a user-defined trajectory.
    }
	\label{fig:MPCFunc}

    \centering
    \includegraphics[width= 0.23\columnwidth]{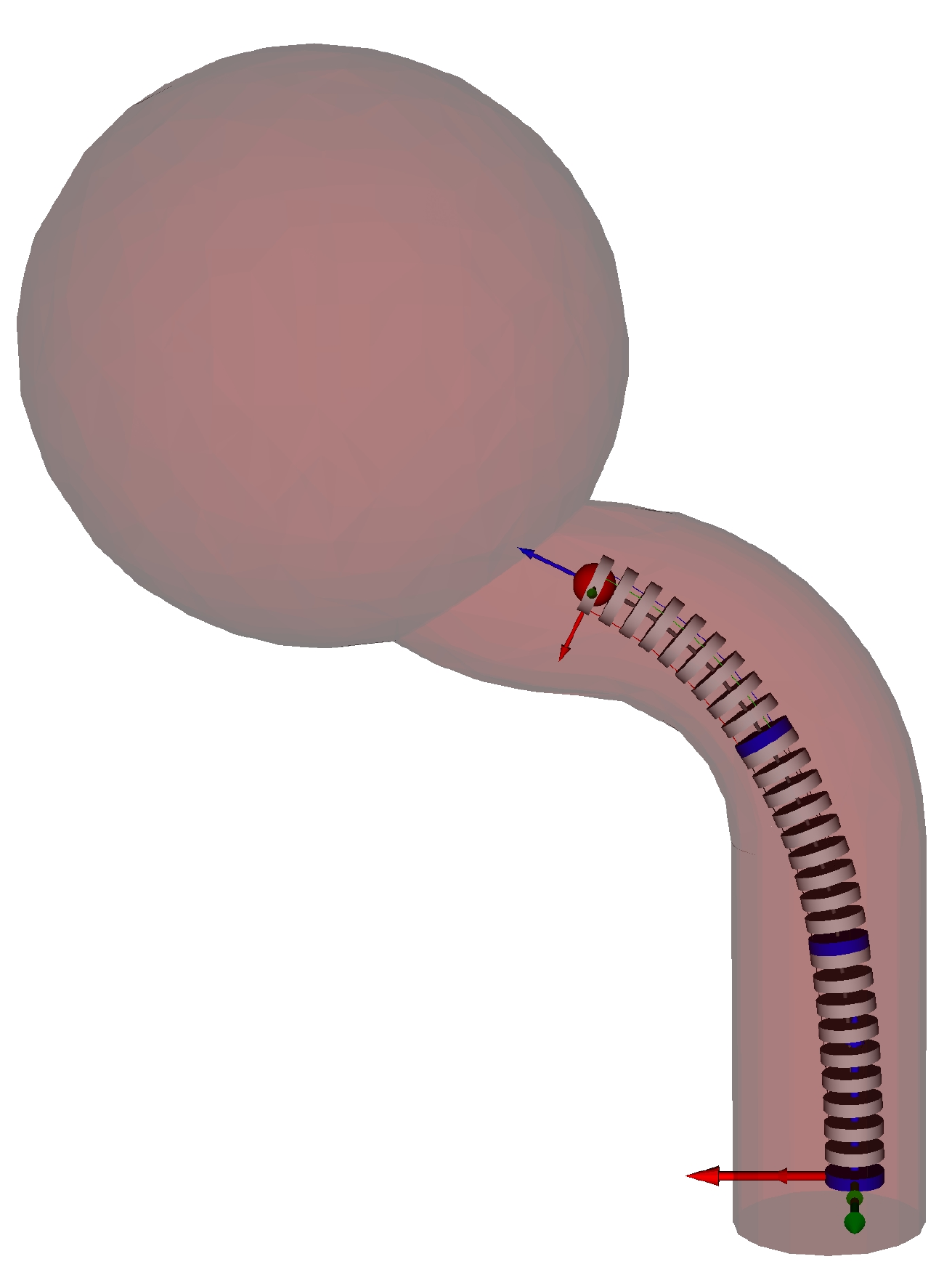}
    \hfill
    \includegraphics[width= 0.23\columnwidth]{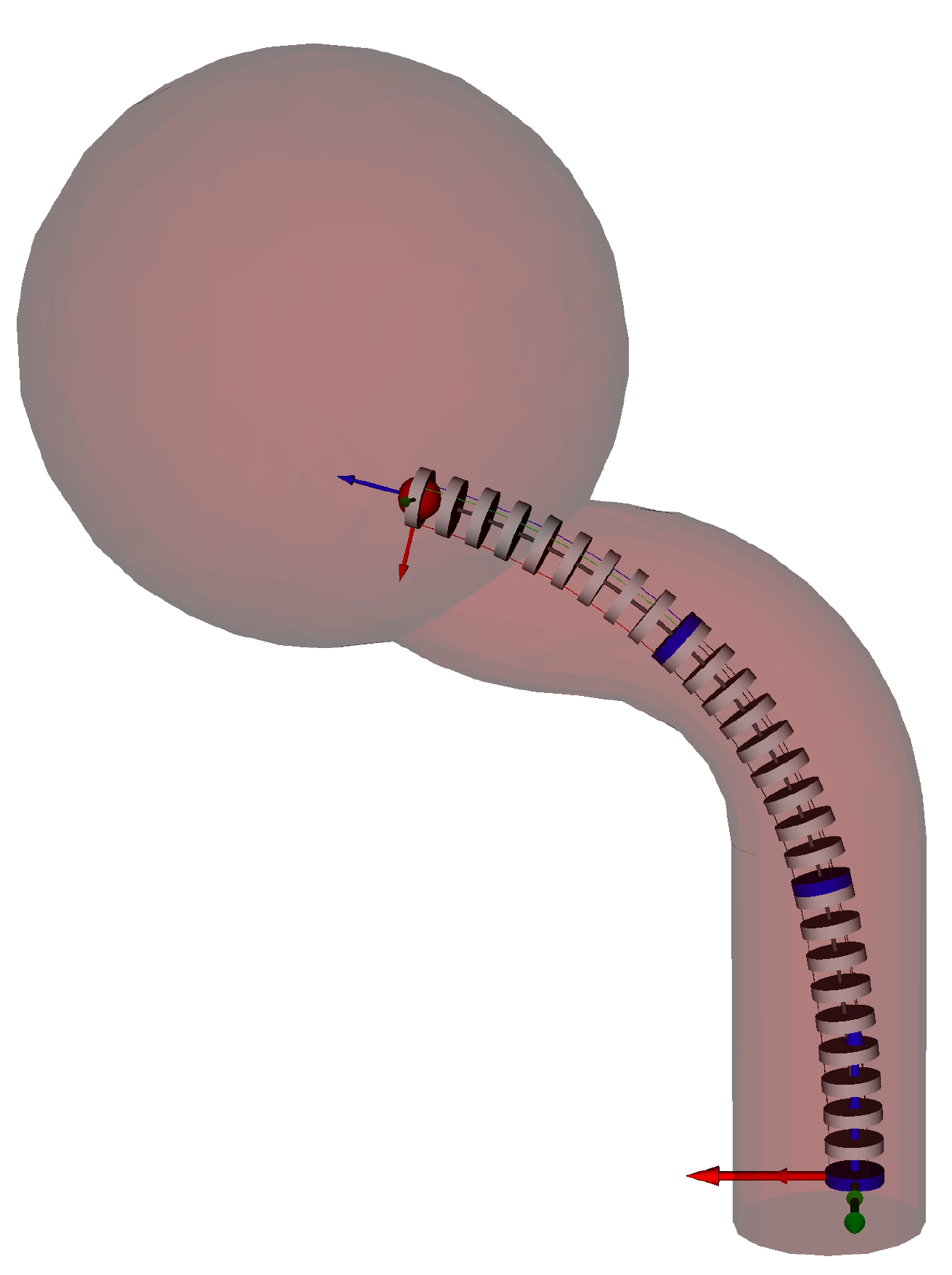}
    \hfill
    \includegraphics[width= 0.23\columnwidth]{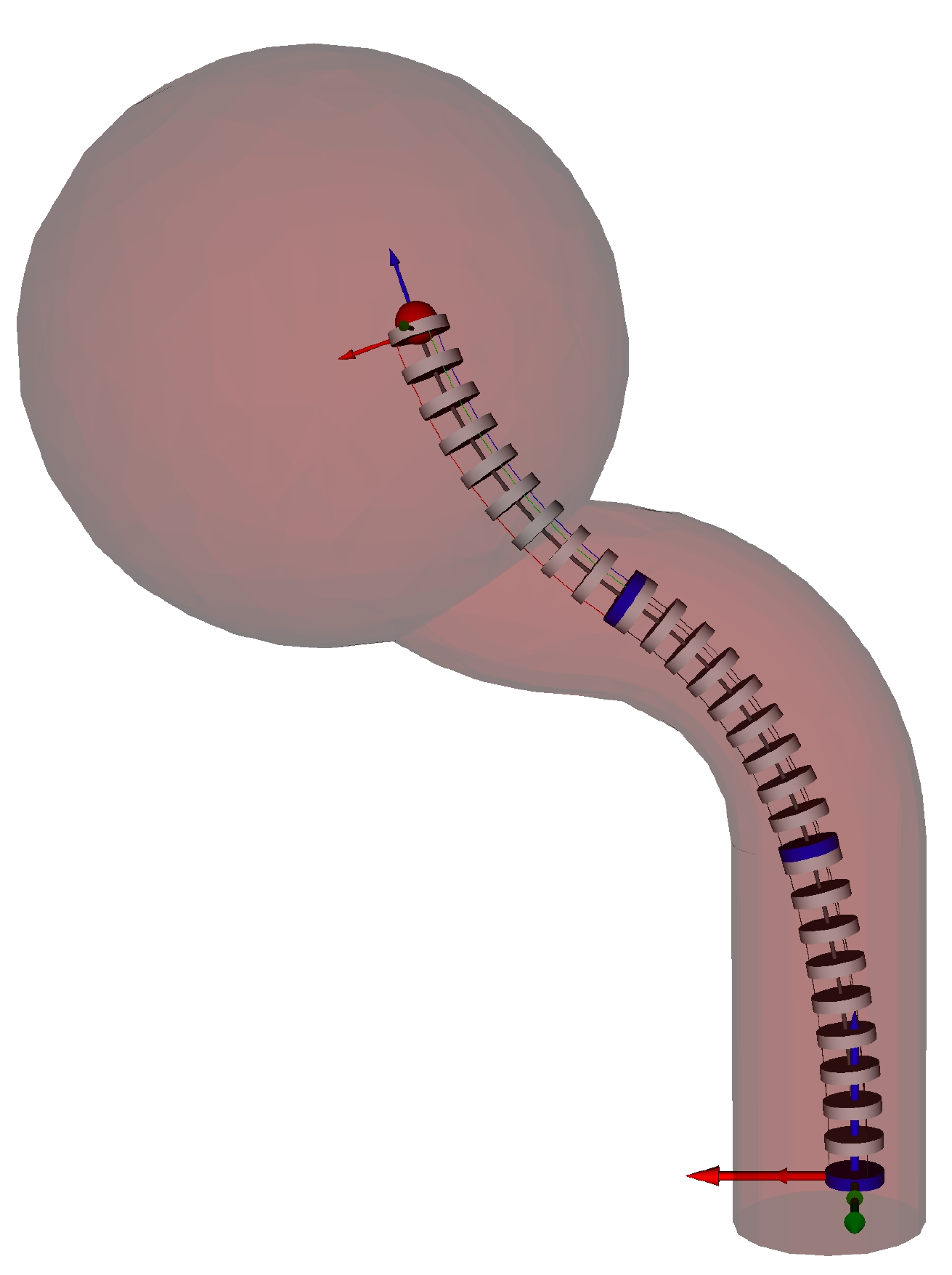}
    \hfill
    \includegraphics[width= 0.23\columnwidth]{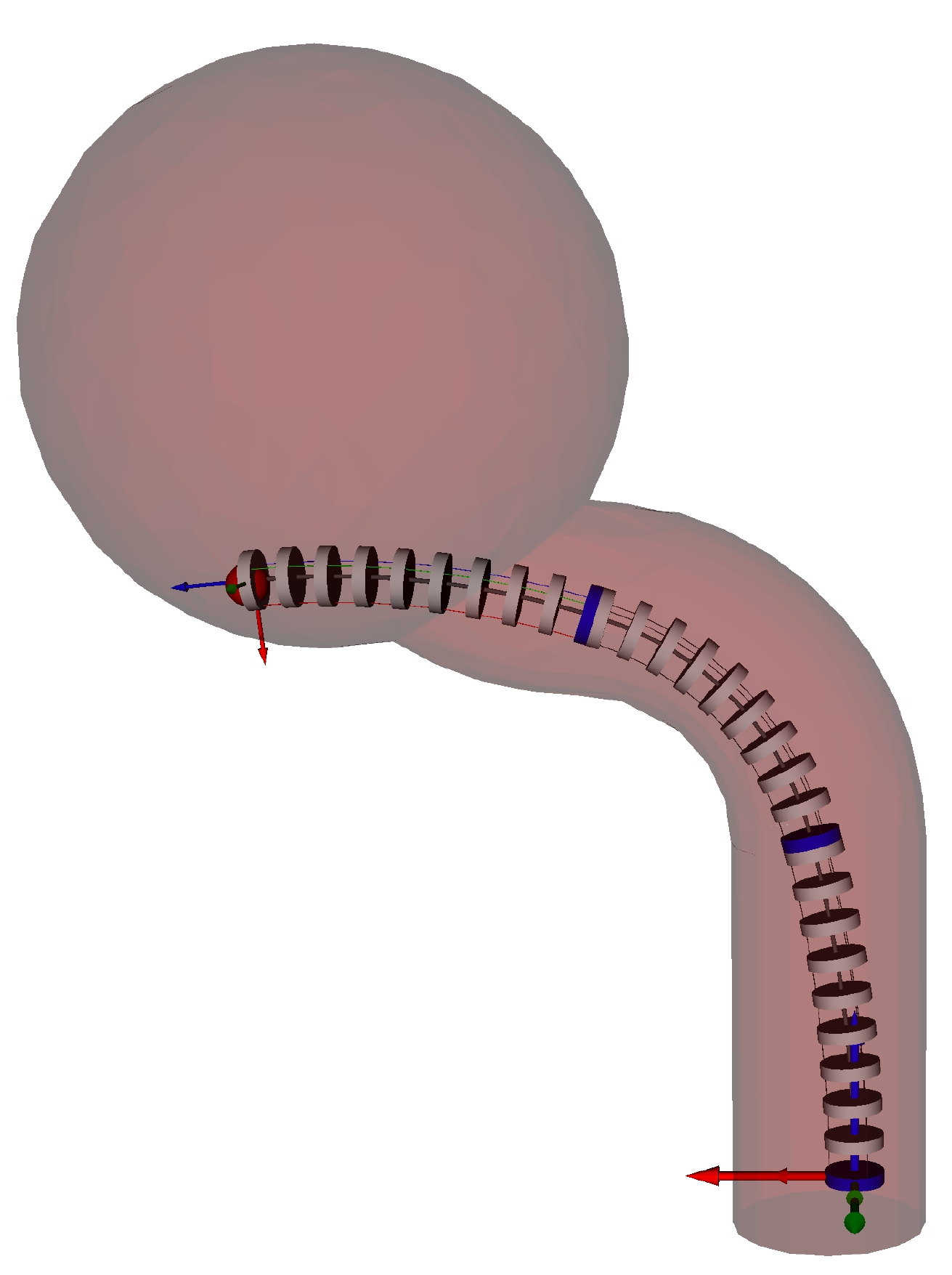}
	\caption{
        Trajectory resulting from application of the DLS controller closed loop steering to four consecutive points describing a user-defined trajectory.
    }
	\label{fig:JacFunc}
\end{figure}

To evaluate the performance of steering the EE while satisfying shape constraints, both controllers are tested to reach four consecutive goal points within a winded tube environment.
The results are shown in Fig.~\ref{fig:MPCFunc} and Fig.~\ref{fig:JacFunc}.

The controllers must reach the goal positions while staying inside the predefined safe zone.
In addition, the controller should steer the shape to a configuration with minimal bending curvatures to avoid the need of reconfiguration at following movements.

\begin{figure}[h]
	\centering
    \includegraphics[width=\columnwidth]{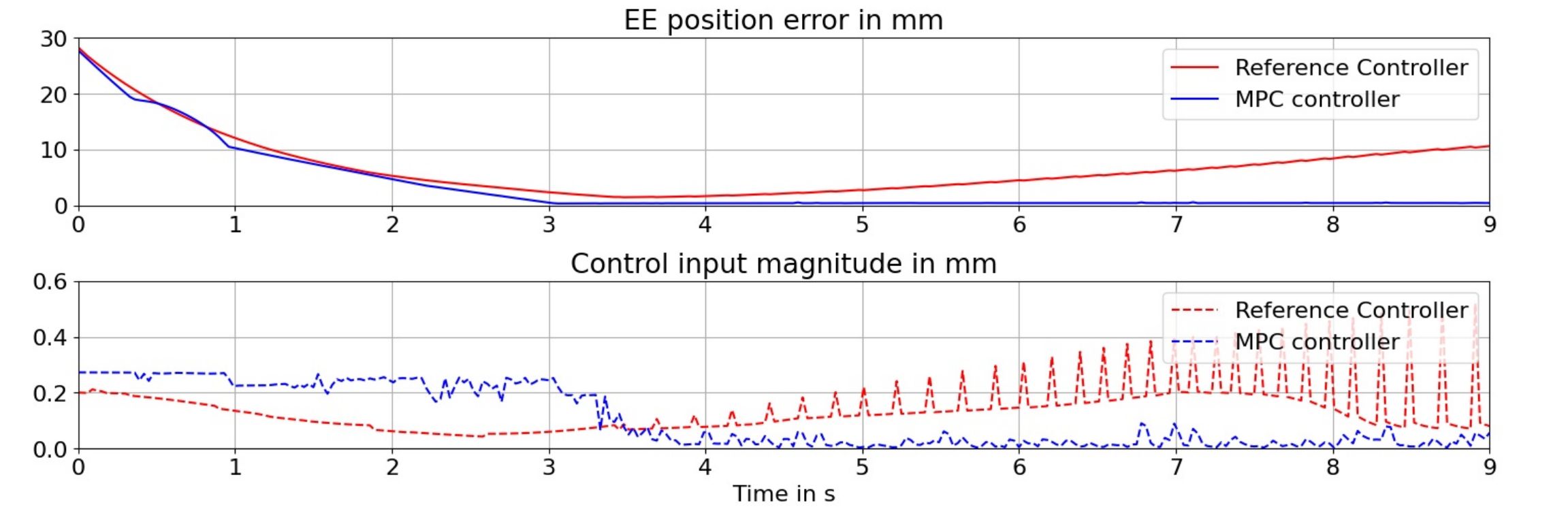}
    \caption{Trajectory of the reference controller becoming unstable while trying to reach a point at the edge of the feasible robot workspace. In comparison, the MPC controller shows stable convergence.}
    \label{fig:RefFail}
\end{figure}

Both controllers showed similar results and reached the goal positions in the desired way.
The tube-shaped part of the hull forced the controllers to perform a follow-the-leader-like motion to extend the robot.
However, the DLS controller encountered stability issues in some configurations, one example is shown in Fig. \ref{fig:RefFail}, especially in configurations at the edge of the reachable robot workspace.
The MPC controller shows stable behaviour in the considered test cases.

We show another example in the first part of the video attachment, where the robot navigates through an inverted U-shaped environment to reach its target.

\subsection{Testing Shape Constraints}
\label{ShapeConstSim}

\begin{figure}[h]
\centering
    \includegraphics[width=0.3\columnwidth]{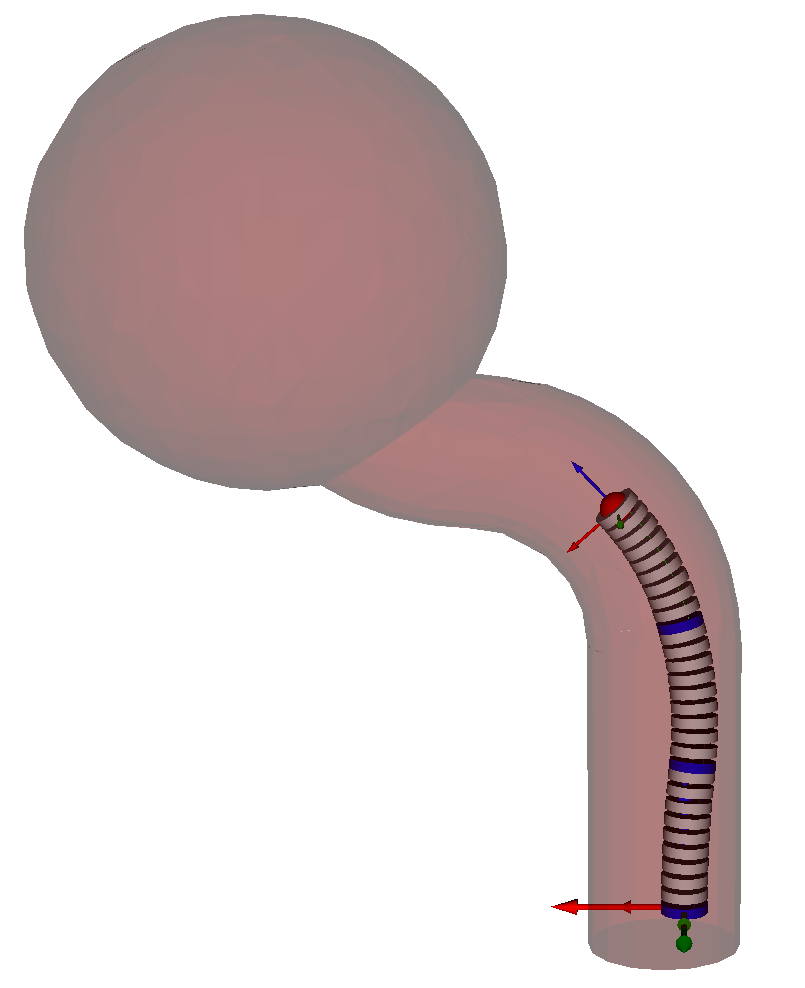}
    \includegraphics[width=0.3\columnwidth]{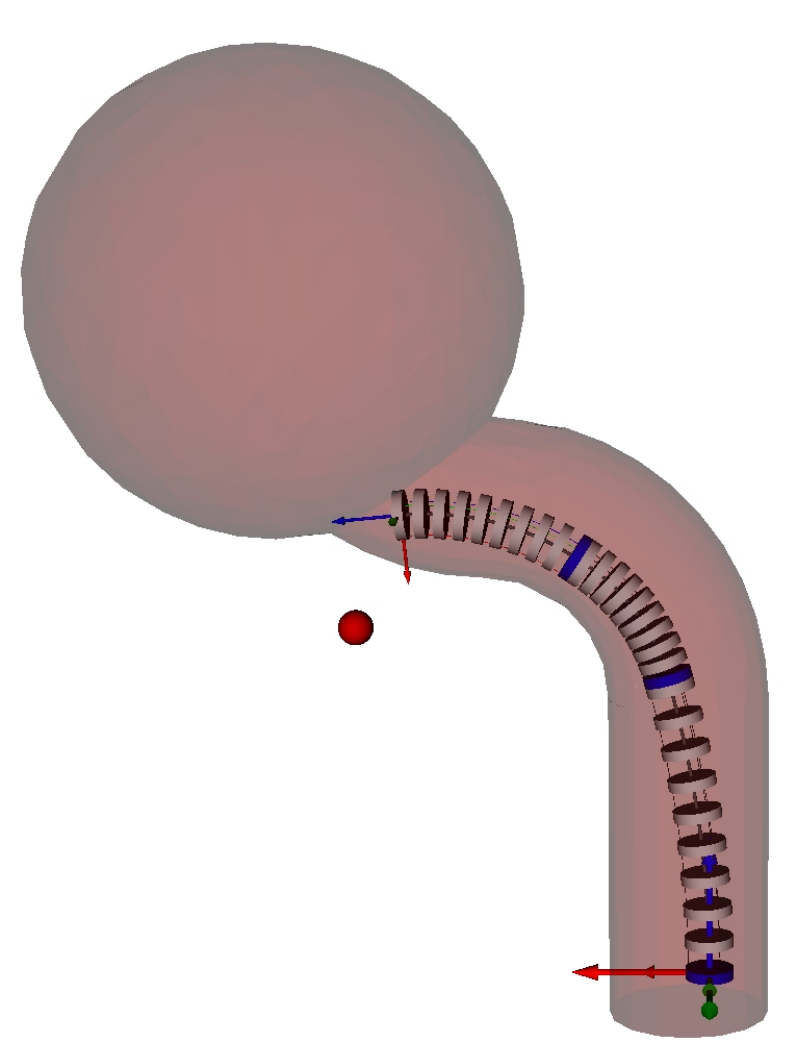}
	\includegraphics[width=0.3\columnwidth]{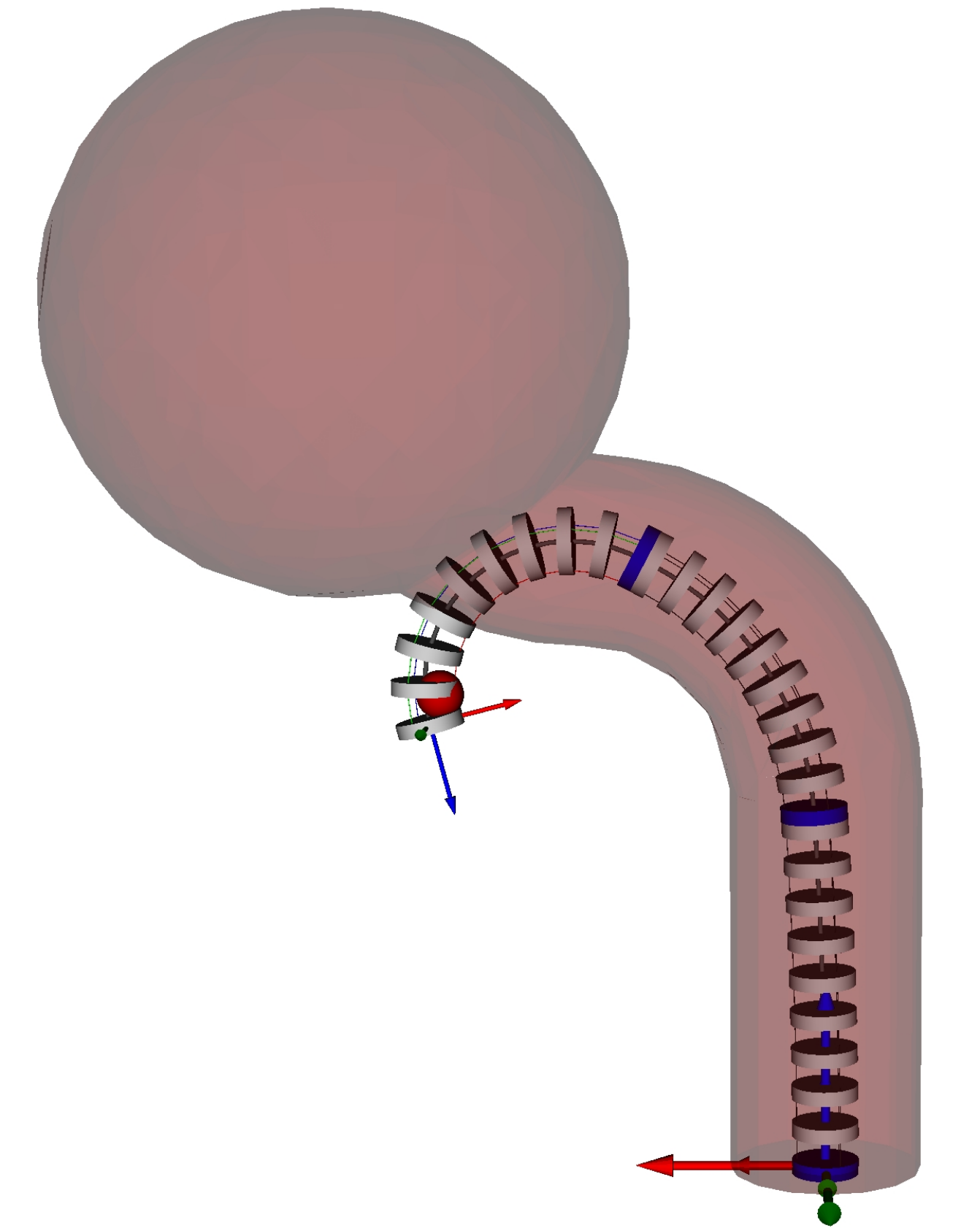}
    \caption{
        Starting from a feasible initial configuration (left), reached steady state configuration of the MPC controller (middle) and the DLS controller (right). $\pvec_d$ is not reachable as it lies outside of the admissible state set.
        }
	\label{fig:Kaput}
\end{figure}

To see if the controllers are able to restrict the robot shape to within the defined safe zone, we define an unreachable $\pvec_d$ outside.
If successful, the robot body should be kept at the desired distance to the safe zone, while the EE should be brought close to the desired position.
The robot configurations and shapes after convergence are depicted in Fig.~\ref{fig:Kaput} for both controllers.

The MPC controller keeps the EE at a desired safety margin to the safe zone while being close to the goal position, since collision avoidance is formulated as a hard constraint. The DLS controller, however, violates the safe zone to reach the goal position. This is inherent to Jacobian-based controllers with collision avoidance formulated as an additive cost function --- they give rise to tuning difficulties where the robot can be overly aggressive or conservative depending on the weighting. 

To mitigate such failure cases for the DLS controller, one may suggest projecting the goal position onto the user-define safe zone. However, besides the additional computation, such projections can be ambiguous or volatile for safe zones with complex geometry and sharp edges, which ultimately leads to unpredictable robot behaviour and unintutive teleoperation experience for the users.

\subsection{Computational Performance}
With a prediction horizon of $N=2$ and the safe-zone defined by $350$ vertices, an average iteration time of \SI{30}{ms} is achieved on an Intel Core i7-9750H CPU with 6 cores at 2.60GHz and 32 GB RAM.
With a higher number of prediction steps, the iteration time increases.
For values of $N < 6$ without shape constraints and $N<4$ with shape constraints, an average iteration time of less than \SI{100}{ms} is attainable.

\section{Evaluation on Real Prototype}

\begin{figure}
	\centering	\includegraphics[width=0.36\columnwidth]{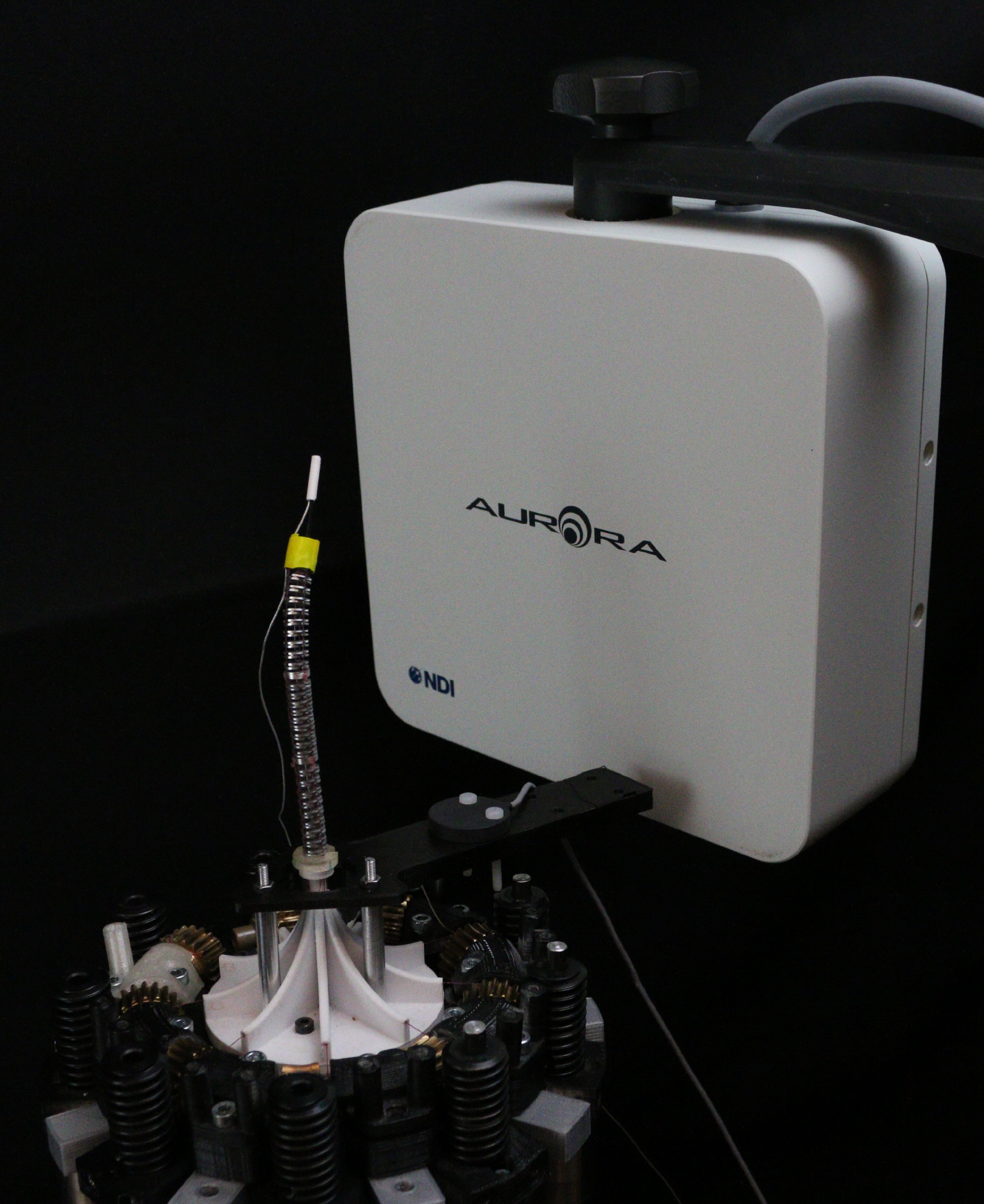}
	\caption{
        Hardware setup to validate the proposed controller.
        Shown is the TDCR prototype with attached measurement coil (white) above the EE (yellow) and the Aurora system in the background.}
	\label{fig:Figmeta}
\end{figure}

For all hardware tests, the robotic prototype presented in \cite{Amanov2021} is used.
At its actuation unit, DC motors drive tendons over a worm drive with a high reduction factor.
For position feedback of the robot, an electromagnetic tracking system is used (Aurora, NDI Inc., \SI{1.4}{mm} stated accuracy).
The tracking coil for the EE position is placed on a small non-magnetic stick approximately \SI{1}{cm} above the EE to minimize magnetic interference.
A full body shape measurement is not possible due to the number of  sensing coils needed and magnetic interference from the magnetic spacer disks.
Measurements are acquired at the rate of \SI{40}{Hz}.
The robot and measurement setup is shown in Fig.~\ref{fig:Figmeta}.

\subsection{Convergence Tests}
\begin{figure}[h]
	\centering
	\includegraphics[width=\columnwidth]{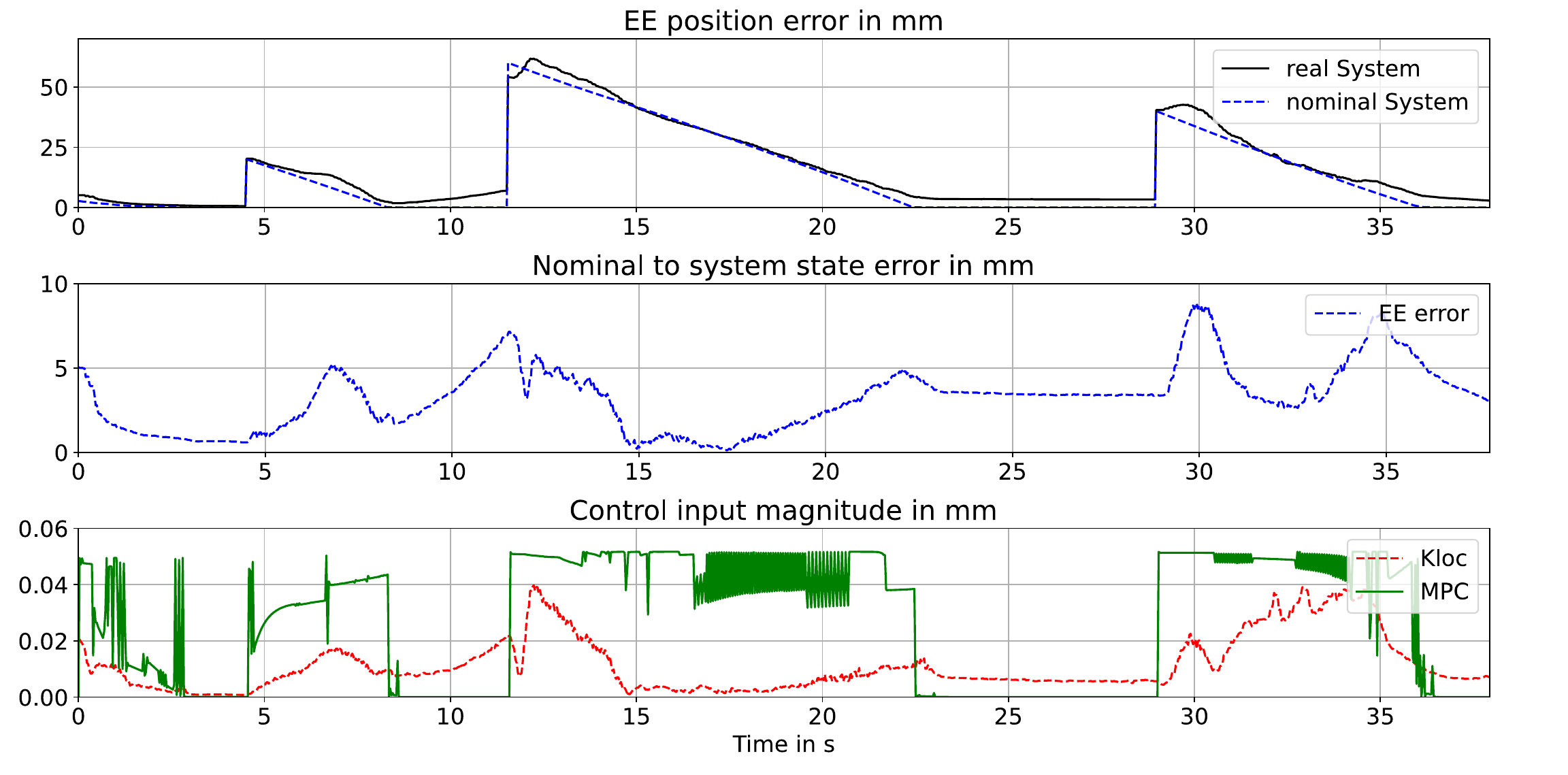}
	\caption{
        Trajectory generated when applying the MPC controller and trying to reach four different goals in the taskspace of the utilized TDCR.
        }
	\label{fig:Meta_traj}
\end{figure}

For testing convergence, the controller had to steer the robot to multiple consecutive positions in task space. The points were chosen in a 2D-plane, spanning the allowed used-defined safe zone, where straight EE trajectories exist in between (see Fig. \ref{fig:catchy} and video attachment). The initial configuration of the robot was a straight configuration with the robot length reduced to about 10 \% of the minimal robot length.
At first, a point relatively far away (\SI{55}{mm}) was set as a goal. 
Three other points were chosen which had a distance to the current EE position of around \SI{30}{mm} and spanned the allowed safe zone for the EE.
The acquired trajectory data is depicted in Fig.~\ref{fig:Meta_traj}.

In the first part of the trajectory (\SI{0}{s} to \SI{17}{s}), the controller shows similar performance as in simulation.
Initially, the nominal to real system state EE error follows a similar dynamic as observed in simulation.
From approximately \SI{17}{s}, the error increases and the local controller is only partly able to compensate the error between nominal state and actual system state.
In the following section of the trajectory, the nominal system has reached its goal but the real system has not.
During the interval from \SI{22}{s} to \SI{29}{s}, the error between the nominal and actual system remains constant, despite a constant input from the local controller.
This behaviour is due to limitation with the prototype: tendons can lose tension at high robot curvatures, as there is no mechanism for tendon tensioning or tension control incorporated into the actuators.
In such cases, the system becomes uncontrollable, and inputs from the local controller may have no effect.
For the tested trajectory, the issue can be resolved by returning the robot to a more straight configuration.

During the test, the robot actuator speed was kept low (\SI{4.5}{mm/s} for tendons and \SI{0.9}{mm/s} for backbone) compared to motor limits to reduce dynamic robot motion.
By this, the modeling assumption that the EE position only depends on static effects was fulfilled.
An additional example real robot evaluation can be seen in the second part of the video attachment.

\subsection{Holding a Shape Constraints}

\begin{figure}[h]
	\centering
	\includegraphics[width=\columnwidth]{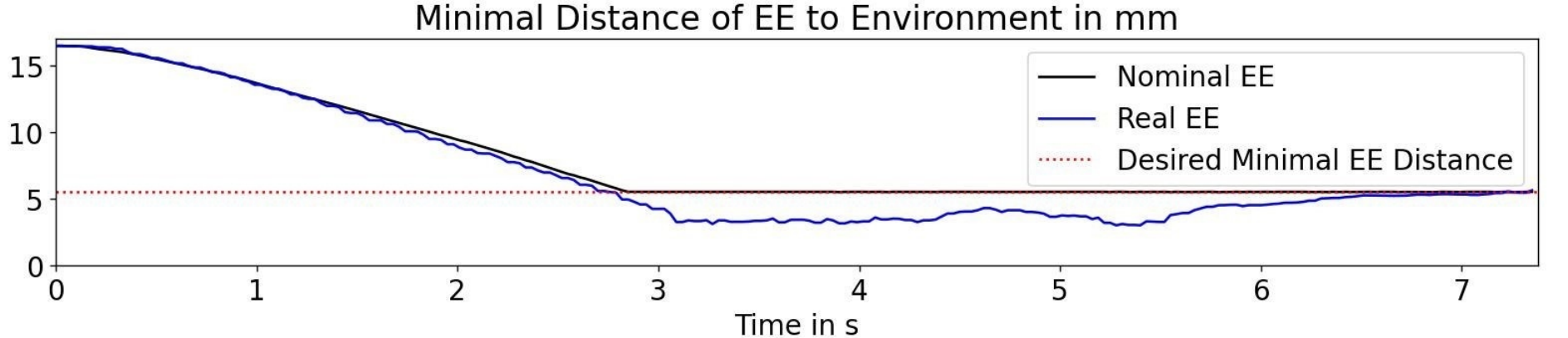}
	\caption{
        The MPC is able to hold the TDCR prototype at the desired minimum distance despite an unfeasible $\pvec_d$ input.
    }
	\label{fig:safetyreal}
\end{figure}

As in simulation, holding shape constraints is also tested on the real system by defining an unfeasible $\pvec_d$ out of bounds of a simple environment.
As the sensor only provides feedback for the EE position, an evaluation is only possible with respect to the robot EE instead of the whole robot body.
The resulting trajectory is depicted in Fig.~\ref{fig:safetyreal}.
The nominal system is accurately holding the constraints.
For the real system, an overshooting movement that reduces the distance to the boundary by a maximum of \SI{2.5}{mm} is observed, as it is influenced by disturbances.
This deviation is still acceptable since a safety margin of \SI{5.5}{mm} was selected to ensure that the distance remains above zero.

\section{Discussion}

The presented results show that the MPC is able to steer a TDCR in confined and complex environments to feasible EE points while ensuring that neither the tip or body of the robot enters undesired areas of the workspace. 
When choosing goal points respecting the environmental constraints, this behavior is also achievable with the compared DLS controller. 
However, the MPC controller brought a distinct advantage by ensuring that the tip is also kept in the safe area for unfeasible targets.
By this, the controller can act as a watchdog in teleoperation applications, as the safe zone is automatically enforced by the controller. This strategy offers several advantages over alternative approaches, such as projecting target positions to the safe zone. It results in fewer unexpected behaviors, particularly in non-convex environments where this projection may change abruptly, thus provides a more intuitive experience for users.

Furthermore, no stability issues occurred during operation which is paramount for safety-critical applications. Unstable behaviour may harm the robot and its surroundings.
The MPC controller allows for straightforward imposition of hard limits on actuator speed and EE speed. Defining the feasible workspace can be easily accomplished using a 3D mesh, which is more challenging with a Jacobian-based approach.
Overall, the MPC controller demonstrated distinct advantages over the reference controller in three key aspects:
\begin{itemize}
    \item Explicit collision avoidance constraints
    \item Stability and convergence under disturbances
    \item Straightforward formulation of control limits
\end{itemize}

The computational performance of the MPC is promising but has to be improved in future research for real-world application.
As we consider an online problem where target positions are given in real-time, offline planning is not a feasible solution.
The used model is computationally efficient, but approximates the robot without considering dynamic effects or external forces.
More sophisticated models that could address these limitations exist but often come with a higher computational cost \cite{Rao15}.

The robot prototype used during experiments has some hardware limitations that compromise control performance. Feedback about the actual tendon tension is not available. Therefore, tendons may become loose and uncontrollable during operation. In the future, precise and fast state estimation of the entire robot body could be integrated, for instance, by using Fiber Bragg strain/shape sensing \cite{Lilge2022}.

Looking beyond teleoperation settings, our proposed controller could be integrated with high-level global planners for autonomous task execution while ensuring safety.

\section{Conclusion}

MPC was used to efficiently control the EE of a TDCR while making sure that the robot's body remains within a user-defined safe zone.
It demonstrates promise in imposing explicit hard constraints on the entire body of the TDCR. Unlike previous methods, these constraints could be general and complex in shape. 
Utilizing a 3D mesh for constraint definition, the formulation of the controller remains straightforward.
To the best of the authors' knowledge, this is the first controller capable of handling general shape constraints and joint velocity limits for a TDCR.

The proposed MPC is also capable of finding an efficient shape within the given bounds without user interaction. Therefore, it not only manages the redundancy of the TDCR but also leverages it as an advantage.

Future work could enhance the controller's robustness against hardware limitations, such as the potential loss of tendon tension. Additionally, incorporating external sensor information through state estimation methods could enhance the controller's performance for sensorized TDCRs.

\section*{Acknowledgment}

The first author was an International Visiting Graduate Student (IVGS) from the Leibniz University Hannover, Germany. 
We thank Matthias M\"uller and Torsten Lilge for supporting the IVGS program.
We thank Reinhard M. Grassmann for discussions aiding the conceptualization of this letter.

\bibliographystyle{IEEEtran}
\bibliography{literature}

\end{document}